\documentclass[10pt,twocolumn,letterpaper]{article}

\usepackage{times}
\usepackage{epsfig}
\usepackage{graphicx}
\usepackage{amsmath}
\usepackage{amssymb}
\usepackage{booktabs}
\usepackage{multirow}
\usepackage{caption}
\usepackage{subcaption}
\usepackage{hyperref}
\usepackage{algorithm}
\usepackage{algorithmic}
\usepackage{xcolor}
\usepackage{enumitem}

\usepackage[letterpaper, margin=1in]{geometry}

\title{Image Classification via Random Dilated Convolution with Multi-Branch Feature Extraction and Context Excitation}

\author{
Wentao Jiang \quad Yuanchan Xu$^{*}$ \quad Heng Yuan \\
Sichuan University
}

\date{}

\begin{document}
\maketitle

\begin{abstract}
Image classification remains a fundamental yet challenging task in computer vision, particularly when fine-grained feature extraction and background noise suppression are required simultaneously. Conventional convolutional neural networks, despite their remarkable success in hierarchical feature learning, often struggle with capturing multi-scale contextual information and are susceptible to overfitting when confronted with noisy or irrelevant image regions. In this paper, we propose RDCNet (Image Classification Network with Random Dilated Convolution), a novel architecture built upon ResNet-34 that integrates three synergistic innovations to address these limitations: (1) a Multi-Branch Random Dilated Convolution (MRDC) module that employs parallel branches with varying dilation rates combined with a stochastic masking mechanism to capture fine-grained features across multiple scales while enhancing robustness against noise and overfitting; (2) a Fine-Grained Feature Enhancement (FGFE) module embedded within MRDC that bridges global contextual information with local feature representations through adaptive pooling and bilinear interpolation, thereby amplifying sensitivity to subtle visual patterns; and (3) a Context Excitation (CE) module that leverages softmax-based spatial attention and channel recalibration to dynamically emphasize task-relevant features while suppressing background interference. Extensive experiments conducted on five benchmark datasets---CIFAR-10, CIFAR-100, SVHN, Imagenette, and Imagewoof---demonstrate that RDCNet consistently achieves state-of-the-art classification accuracy, outperforming the second-best competing methods by margins of 0.02\%, 1.12\%, 0.18\%, 4.73\%, and 3.56\%, respectively, thereby validating the effectiveness and generalizability of the proposed approach across diverse visual recognition scenarios.
\end{abstract}

\section{Introduction}

Image classification constitutes one of the most fundamental problems in computer vision, serving as a critical building block for numerous downstream tasks such as object detection~\cite{tang2022few}, scene understanding~\cite{feng2023docpedia}, and visual question answering~\cite{tang2025mtvqa}. The objective is to assign a given image to one of several predefined categories based on its visual content. With the advent of deep learning, convolutional neural networks (CNNs) have revolutionized this domain, achieving unprecedented performance on large-scale benchmarks~\cite{krizhevsky2017imagenet,he2016deep}.

The evolution of CNN architectures for image classification has progressed through several pivotal milestones. LeNet~\cite{lecun1998gradient} pioneered the use of multi-layer convolutional and pooling operations to extract hierarchical feature representations from raw images, substantially reducing the need for manual feature engineering. AlexNet~\cite{krizhevsky2017imagenet} subsequently demonstrated that increasing network depth enables the learning of more complex and abstract feature hierarchies, catalyzing the deep learning revolution. VGGNet~\cite{simonyan2015very} and GoogLeNet~\cite{szegedy2015going} further validated the advantages of deeper architectures, with the Inception module in GoogLeNet achieving multi-scale feature extraction through parallel convolutions of different kernel sizes. However, these deeper networks also exposed critical challenges including vanishing and exploding gradients during training.

The introduction of residual connections in ResNet~\cite{he2016deep} represented a watershed moment, effectively alleviating gradient degradation in very deep networks through identity shortcut connections. Building upon this foundation, Wide Residual Networks (WRN)~\cite{zagoruyko2017wide} explored the benefits of increasing channel width rather than depth, DenseNet~\cite{huang2017densely} employed dense connectivity patterns to facilitate feature reuse and gradient flow, and MobileNet~\cite{howard2017mobilenets} introduced depthwise separable convolutions for lightweight deployment. More recently, vision transformers~\cite{dosovitskiy2021image,liu2021swin} have demonstrated the power of self-attention mechanisms for capturing long-range dependencies in visual data. Despite these advances, several fundamental challenges persist in image classification that motivate the present work.

\textbf{Challenge 1: Fine-grained feature extraction.} Standard convolutional operations with fixed kernel sizes primarily capture local patterns within a constrained receptive field. While stacking multiple convolutional layers gradually expands the effective receptive field, this approach is computationally expensive and may still fail to capture subtle discriminative features at multiple spatial scales simultaneously. This limitation is particularly pronounced when distinguishing between visually similar categories that differ only in fine-grained details, such as breeds of dogs or species of birds. Multi-resolution approaches such as HRNet~\cite{wang2022dual} and Multi-ResNet~\cite{abdi2017multi} have attempted to address this issue but introduce significant computational overhead. Similarly, HO-ResNet~\cite{luo2022rethinking} enhances feature extraction through high-order feature modeling but lacks precise spatial localization of discriminative regions.

\textbf{Challenge 2: Background noise and irrelevant region interference.} In real-world images, the target object often occupies only a fraction of the visual field, with the remainder consisting of cluttered backgrounds, occlusions, and other distracting elements. Conventional CNNs process all spatial regions with equal computational effort, making them vulnerable to background noise that can mislead the classification decision. Attention mechanisms such as the Squeeze-and-Excitation (SE) module~\cite{hu2018squeeze}, CBAM~\cite{woo2018cbam}, and Non-Local blocks~\cite{wang2018non} have been proposed to selectively enhance informative features while suppressing irrelevant ones. However, SE lacks spatial awareness, Non-Local modules incur prohibitive computational costs due to dense pairwise similarity computations, and CCNet~\cite{huang2019ccnet} is limited in capturing full global dependencies due to its criss-cross attention pattern.

\textbf{Challenge 3: Overfitting and generalization.} Deep neural networks with millions of parameters are inherently prone to overfitting, particularly when training data is limited or noisy. Data augmentation techniques such as random erasing~\cite{zhong2020random}, Cutout~\cite{devries2017improved}, CutMix~\cite{yun2019cutmix}, and mixup~\cite{zhang2018mixup} have been widely adopted to mitigate this issue at the data level. At the architectural level, Dropout~\cite{srivastava2014dropout} randomly deactivates neurons during training to prevent co-adaptation. However, these strategies operate at relatively coarse granularities and do not directly address the structural limitations of convolutional operations.

Dilated (atrous) convolutions~\cite{yu2016multi,chen2018deeplab} offer an elegant solution to the receptive field limitation by inserting gaps between kernel elements, effectively expanding the field of view without increasing the number of parameters. The Atrous Spatial Pyramid Pooling (ASPP) module in DeepLab~\cite{chen2018deeplab,chen2017rethinking} employs parallel dilated convolutions at multiple rates to capture multi-scale contextual information. However, the fixed dilation patterns in these approaches can lead to gridding artifacts~\cite{yu2016multi}, where certain spatial positions are systematically ignored, resulting in information loss at fine-grained levels. Frequency-adaptive dilated convolution (FADC)~\cite{chen2024frequency} and SFHFormer~\cite{jiang2025when} have attempted to enhance fine-grained feature extraction but do not fully resolve the fundamental tension between receptive field expansion and local detail preservation.

To address these interrelated challenges, we propose RDCNet (Image Classification Network with Random Dilated Convolution), a novel architecture that integrates several complementary innovations within a unified framework built upon ResNet-34~\cite{he2016deep}. Our key contributions are as follows:

\begin{enumerate}[leftmargin=*]
\item We propose the \textbf{Multi-Branch Random Dilated Convolution (MRDC)} module, which replaces the standard $3\times3$ convolution in the BasicBlock with a multi-branch structure incorporating dilated convolutions at multiple rates. By coupling dilated convolutions with a novel \textbf{random masking mechanism} that stochastically masks both input channels and kernel weights, MRDC simultaneously expands the receptive field, mitigates the gridding effect, and provides built-in regularization that enhances robustness.

\item We introduce the \textbf{Fine-Grained Feature Enhancement (FGFE)} module as a dedicated branch within MRDC that bridges global and local feature representations. By applying global average pooling followed by $1\times1$ convolution and bilinear upsampling, FGFE transforms global contextual information into spatially-aware local feature enhancements, improving the network's sensitivity to subtle discriminative patterns.

\item We propose the \textbf{Context Excitation (CE)} module, which employs a softmax-based spatial attention mechanism combined with channel recalibration to dynamically adjust feature importance. Unlike SE~\cite{hu2018squeeze}, CE captures both spatial and channel dependencies; unlike Non-Local~\cite{wang2018non}, it achieves this with substantially lower computational cost through an efficient key-value mechanism.

\item Comprehensive experiments on five diverse benchmarks demonstrate that RDCNet achieves consistent state-of-the-art performance, with ablation studies confirming the complementary contributions of each proposed module.
\end{enumerate}

\section{Related Work}

\subsection{Convolutional Neural Networks for Image Classification}

The trajectory of CNN-based image classification has been marked by increasingly sophisticated architectural designs. Early architectures such as LeNet~\cite{lecun1998gradient} and AlexNet~\cite{krizhevsky2017imagenet} established the fundamental paradigm of learning hierarchical feature representations through stacked convolutional layers. VGGNet~\cite{simonyan2015very} demonstrated that using small $3\times3$ kernels throughout the network achieves superior performance while reducing the parameter count compared to larger kernels. GoogLeNet~\cite{szegedy2015going} introduced the Inception module for multi-scale feature extraction within a single layer, while ResNet~\cite{he2016deep} solved the degradation problem in very deep networks through residual learning. Subsequent works have explored various dimensions of architectural design, including wider networks (WRN~\cite{zagoruyko2017wide}), dense connectivity (DenseNet~\cite{huang2017densely}), depthwise separable convolutions (MobileNet~\cite{howard2017mobilenets}, MobileNetV2~\cite{sandler2018mobilenetv2}), channel shuffling (ShuffleNetV2~\cite{ma2018shufflenet}), neural architecture search (EfficientNet~\cite{tan2020efficientnet}), and ghost feature generation (GhostNet~\cite{han2020ghostnet}). More recently, vision transformers~\cite{dosovitskiy2021image,liu2021swin} and hybrid architectures such as Couplformer~\cite{lan2021couplformer} and FAVOR+~\cite{choromanski2020rethinking} have emerged as powerful alternatives, though pure CNN designs remain competitive in many practical scenarios due to their inductive biases and computational efficiency.

In the context of fine-grained and robust image classification, several recent methods are particularly relevant. MMA-CCT~\cite{konstantinidis2023multi} combines multi-manifold attention with cascaded cross-scale transformers to enhance feature discrimination. ATONet~\cite{wu2024auto} employs automatic network pruning guided by a controller network to achieve efficient classification. QKFormer~\cite{zhou2024qkformer} adapts spiking transformers with Q-K attention for energy-efficient visual recognition. DAMSNet~\cite{jiang2023double} utilizes double-branch multi-attention mechanisms with sharpness-aware optimization, while SSCNet~\cite{jiang2024sparse} introduces spatial position correction for sparse feature classification. These methods demonstrate the ongoing need for improved feature extraction mechanisms in image classification, which our proposed RDCNet addresses through a novel combination of randomized dilated convolutions and context-aware feature enhancement.

\subsection{Dilated Convolution and Multi-Scale Feature Extraction}

Dilated (atrous) convolutions were first introduced by Yu and Koltun~\cite{yu2016multi} for dense prediction tasks, enabling exponential expansion of the receptive field without increasing the number of parameters or reducing spatial resolution. This concept was further developed in DeepLab~\cite{chen2018deeplab} and its successors~\cite{chen2017rethinking}, where the ASPP module employs parallel dilated convolutions at multiple rates to capture multi-scale contextual information. The key advantage of dilated convolutions over standard convolutions is their ability to aggregate information from a wider spatial extent while maintaining the same computational complexity.

However, a well-known limitation of dilated convolutions is the gridding effect~\cite{yu2016multi}, where the systematic spacing of kernel elements creates blind spots in the feature sampling pattern. This issue becomes more pronounced at larger dilation rates, potentially leading to the loss of fine-grained spatial details. Various solutions have been proposed, including hybrid dilated convolution (HDC) patterns that alternate different dilation rates~\cite{chen2017rethinking}, deformable convolutions~\cite{dai2017deformable} that learn adaptive sampling positions, and frequency-adaptive dilated convolutions~\cite{chen2024frequency} that dynamically adjust dilation based on frequency content. RFBNet~\cite{liu2018receptive} combines different kernel sizes with dilated convolutions to simulate the receptive field structure of the human visual system, achieving improved multi-scale feature capture but with limited flexibility in adapting to diverse input patterns.

Our proposed MRDC module distinguishes itself from these approaches through the integration of a random masking mechanism with dilated convolutions. Rather than relying on fixed or learned dilation patterns, MRDC stochastically masks both input channels and kernel weights during training, effectively creating a diverse ensemble of feature extraction patterns that mitigate the gridding effect while providing built-in regularization. This approach is conceptually related to Dropout~\cite{srivastava2014dropout} but operates at the level of convolutional operations rather than individual neurons, providing finer-grained control over the regularization effect.

\subsection{Attention Mechanisms and Feature Recalibration}

Attention mechanisms have become indispensable components in modern deep learning architectures, enabling networks to selectively focus on informative features while suppressing irrelevant ones. The Squeeze-and-Excitation (SE) network~\cite{hu2018squeeze} pioneered channel attention by adaptively recalibrating channel-wise feature responses through global average pooling and fully connected layers. While highly effective and computationally efficient, SE operates solely in the channel dimension and lacks spatial awareness.

The Non-Local module~\cite{wang2018non}, inspired by the self-attention mechanism in transformers~\cite{vaswani2017attention}, captures long-range spatial dependencies by computing pairwise similarities between all spatial positions. While powerful, its quadratic computational complexity with respect to spatial resolution limits its applicability in high-resolution feature maps. Subsequent works have sought to reduce this computational burden: CCNet~\cite{huang2019ccnet} restricts attention computation to criss-cross paths, while ANN~\cite{zhu2019asymmetric} employs asymmetric attention to reduce the number of pairwise computations. GCNet~\cite{cao2019gcnet} unifies the strengths of Non-Local and SE by computing a single global attention map shared across all positions, achieving competitive performance with greatly reduced cost.

CBAM~\cite{woo2018cbam} addresses both channel and spatial attention sequentially, applying channel attention followed by spatial attention refinement. However, this sequential approach may not fully exploit the interdependencies between channel and spatial dimensions. Our proposed CE module takes a different approach by computing spatial attention through a softmax-normalized key generation process and subsequently performing channel recalibration through matrix multiplication and sigmoid-gated feature modulation. This design captures both spatial inter-dependencies and channel-wise importance in an integrated manner, achieving a favorable balance between computational efficiency and representational capacity.

The integration of attention mechanisms with convolutional feature extraction has also been explored in the context of document understanding and visual text analysis. Recent large multimodal models~\cite{feng2023unidoc,tang2024textsquare,zhao2024tabpedia} have demonstrated the effectiveness of attention-based feature selection for text-centric visual tasks. Similarly, scene text detection methods~\cite{tang2022few,tang2022optimal,zhao2023multi} have shown that combining local convolutional features with global attention can significantly improve recognition accuracy in cluttered visual environments, providing additional motivation for our attention-enhanced classification framework.

\section{Proposed Method}

In this section, we present the detailed architecture of RDCNet, which is built upon ResNet-34~\cite{he2016deep} and incorporates three novel modules: the Multi-Branch Random Dilated Convolution (MRDC) module, the Fine-Grained Feature Enhancement (FGFE) module, and the Context Excitation (CE) module. The overall architecture of RDCNet is illustrated in Figure~1 and consists of five sequential stages: preprocessing, initial feature extraction, deep feature extraction with MRDC residual blocks, global feature integration with the CE module, and classification.

\subsection{Shallow Feature Extraction}

The initial feature extraction stage in the original ResNet-34 employs a $7\times7$ convolutional layer followed by a $3\times3$ max pooling layer for primary feature extraction. While effective for large-resolution inputs, this configuration can be overly aggressive for smaller images such as those in CIFAR-10 ($32\times32$ pixels), where the large kernel and pooling operation dramatically reduce spatial resolution and discard fine-grained details in the early layers.

To address this issue, RDCNet adaptively adjusts the initial feature extraction based on the input resolution. For small-resolution datasets (CIFAR-10, CIFAR-100, SVHN with $32\times32$ inputs), we replace the $7\times7$ convolution and max pooling with a single $3\times3$ convolution with stride 1 and padding 1:
\begin{equation}
W_{\text{out}} = \frac{W_{\text{in}} + 2p - k}{s} + 1
\label{eq:conv_size}
\end{equation}
where $W_{\text{in}}$ and $W_{\text{out}}$ are the input and output feature map dimensions, $p$ is the padding size, $k$ is the kernel size, and $s$ is the stride. With $k=3$, $s=1$, and $p=1$, the spatial resolution is preserved at $32\times32$, retaining critical fine-grained information for subsequent processing. For larger-resolution datasets (Imagenette, Imagewoof with $224\times224$ inputs), the standard $7\times7$ convolution and max pooling configuration is retained to provide appropriate dimensionality reduction.

\subsection{Multi-Branch Random Dilated Convolution (MRDC)}

The MRDC module is the core innovation of RDCNet, designed to extract multi-scale fine-grained features through a combination of parallel processing branches with diverse receptive fields, augmented by a stochastic masking mechanism for enhanced robustness. It replaces the second $3\times3$ convolution in the standard BasicBlock of ResNet-34.

\subsubsection{Dilated Convolution with Random Masking}

Dilated convolution extends the standard convolution by introducing a dilation rate $d$ that controls the spacing between kernel elements. For a kernel of size $k\times k$ with dilation rate $d$, the effective receptive field expands to $(d(k-1)+1) \times (d(k-1)+1)$ without increasing the number of parameters. The output spatial size is given by:
\begin{equation}
w_{\text{out}} = \frac{w_{\text{in}} + 2p - d(k-1) - 1}{s} + 1
\label{eq:dilated_size}
\end{equation}
where $d$ denotes the dilation rate. To maintain spatial resolution, we set the padding $p$ equal to the dilation rate $d$, ensuring that $w_{\text{out}} = w_{\text{in}}$ when $s=1$.

While dilated convolutions effectively enlarge the receptive field, their sparse sampling pattern can introduce gridding artifacts~\cite{yu2016multi}, causing the loss of fine-grained spatial information. To mitigate this issue and simultaneously provide regularization, we introduce a \textbf{random masking mechanism} that operates at two complementary levels: channel masking and kernel masking.

\textbf{Channel masking.} Given an input feature map $\mathbf{X} \in \mathbb{R}^{C \times H \times W}$, we generate a binary channel mask $\mathbf{M}_c \in \{0,1\}^{C \times 1 \times 1}$ where each element is independently sampled:
\begin{equation}
m_c = \begin{cases} 1, & \text{if } r_c < \tau \\ 0, & \text{if } r_c \geq \tau \end{cases}
\label{eq:channel_mask}
\end{equation}
where $r_c \sim \text{Uniform}(0,1)$ is a random sample and $\tau \in [0,1]$ is the retention probability threshold. The masked feature map is obtained through element-wise multiplication:
\begin{equation}
\mathbf{X}_c = \mathbf{X} \otimes \mathbf{M}_c
\label{eq:channel_mask_apply}
\end{equation}

\textbf{Kernel masking.} Additionally, we generate a three-dimensional kernel mask $\mathbf{M}_k \in \{0,1\}^{C \times H \times W}$ with elements independently sampled in the same manner:
\begin{equation}
m_{c,h,w} = \begin{cases} 1, & \text{if } r_{c,h,w} < \tau \\ 0, & \text{if } r_{c,h,w} \geq \tau \end{cases}
\label{eq:kernel_mask}
\end{equation}
The kernel-masked feature map is then:
\begin{equation}
\mathbf{X}_k = \mathbf{X}_c \otimes \mathbf{M}_k
\label{eq:kernel_mask_apply}
\end{equation}

The dual masking mechanism serves multiple purposes. First, by randomly suppressing channels and spatial positions, it disrupts the regular spacing pattern of dilated convolutions, effectively mitigating the gridding effect. Second, the stochastic nature of the masking forces the network to learn robust feature representations that are not overly dependent on any particular subset of inputs or kernel elements, functioning as an implicit regularizer. Third, the element-wise nature of the masking operation incurs negligible additional computational overhead. We set $p=d$ to ensure that edge regions of the feature map are properly covered even under the combined effect of dilation and masking.

\subsubsection{Fine-Grained Feature Enhancement (FGFE) Module}

The FGFE module constitutes the fourth branch of MRDC and is specifically designed to bridge global and local feature representations. Given the branch input $\mathbf{F}_4 \in \mathbb{R}^{c \times h \times w}$, the module first applies Global Average Pooling (GAP) to compress the spatial dimensions and obtain a channel descriptor:
\begin{equation}
\mathbf{F}_p = \text{GAP}(\mathbf{F}_4) = \frac{1}{H \times W} \sum_{h=1}^{H} \sum_{w=1}^{W} F_{c,h,w}
\label{eq:gap}
\end{equation}
yielding $\mathbf{F}_p \in \mathbb{R}^{c \times 1 \times 1}$ that encodes the global information for each channel.

This global descriptor is then transformed through a $1\times1$ convolution followed by batch normalization and ReLU activation to map global information to a local feature representation:
\begin{equation}
\mathbf{F}_l = \text{ReLU}(\text{BN}(\text{Conv}_{1\times1}(\mathbf{F}_p)))
\label{eq:fgfe_transform}
\end{equation}

Finally, the transformed feature map is upsampled back to the original spatial resolution through bilinear interpolation:
\begin{equation}
\mathbf{F}'_4 = \delta(\mathbf{F}_l)
\label{eq:bilinear}
\end{equation}
where $\delta(\cdot)$ denotes the bilinear interpolation operation. The bilinear interpolation computes the target pixel value $f(P)$ from the four nearest neighbors $\{Q_{11}, Q_{12}, Q_{21}, Q_{22}\}$ using successive linear interpolation in horizontal and vertical directions:
\begin{equation}
f(P) \approx \frac{y_2 - y}{y_2 - y_1} f(R_1) + \frac{y - y_1}{y_2 - y_1} f(R_2)
\label{eq:bilinear_detail}
\end{equation}
where $f(R_1)$ and $f(R_2)$ are intermediate horizontal interpolation results.

By converting global contextual information into spatially-aware local feature enhancements, FGFE enables the network to amplify subtle discriminative patterns that might otherwise be overshadowed by dominant features, significantly improving fine-grained classification accuracy.

\subsubsection{MRDC Module Architecture}

The overall MRDC module architecture operates by first splitting the input feature map $\mathbf{F}_{\text{in}} \in \mathbb{R}^{c \times h \times w}$ into four groups along the channel dimension:
\begin{equation}
\mathbf{F}_{\text{in}} = [\mathbf{F}_1, \mathbf{F}_2, \mathbf{F}_3, \mathbf{F}_4]
\label{eq:channel_split}
\end{equation}
where each group has $c/4$ channels. Each group is processed by a dedicated branch:

\textbf{Branch 1:} A standard $3\times3$ convolution that preserves local feature extraction capability and ensures training stability:
\begin{equation}
\mathbf{F}'_1 = \text{Conv}_{3\times3}(\mathbf{F}_1)
\label{eq:branch1}
\end{equation}

\textbf{Branch 2:} Random-masked dilated convolution with dilation rate $d=2$, followed by a $3\times3$ convolution for feature refinement:
\begin{equation}
\mathbf{F}'_2 = \text{Conv}_{3\times3}(\text{M\_DConv}(\mathbf{F}_2))
\label{eq:branch2}
\end{equation}
where $\text{M\_DConv}(\cdot)$ denotes the random-masked dilated convolution operation.

\textbf{Branch 3:} Random-masked dilated convolution with dilation rate $d=3$, followed by asymmetric $1\times5$ and $5\times1$ convolutions to capture directional features efficiently:
\begin{equation}
\mathbf{F}'_3 = \text{Conv}_{5\times1}(\text{Conv}_{1\times5}(\text{M\_DConv}(\mathbf{F}_3)))
\label{eq:branch3}
\end{equation}

\textbf{Branch 4:} The FGFE module (described above):
\begin{equation}
\mathbf{F}'_4 = \text{FGFE}(\mathbf{F}_4)
\label{eq:branch4}
\end{equation}

The outputs of all four branches are concatenated along the channel dimension and then fused through a $1\times1$ convolution to restore the original channel count. The final output incorporates a scaled residual connection:
\begin{equation}
\mathbf{F}_{\text{out}} = \alpha \cdot \text{BN}(\text{Conv}_{1\times1}(\text{Concat}(\mathbf{F}'_1, \mathbf{F}'_2, \mathbf{F}'_3, \mathbf{F}'_4))) + \mathbf{F}_{\text{in}}
\label{eq:mrdc_output}
\end{equation}
where $\alpha$ is a learnable scaling factor that controls the relative contribution of the multi-branch features versus the residual input. As demonstrated in our experiments, the optimal value of $\alpha$ is approximately 0.5, providing a balanced integration of the original and enhanced features.

\subsection{Context Excitation (CE) Module}

Inspired by the self-attention mechanism~\cite{vaswani2017attention} and its successful applications in both natural language processing and visual recognition~\cite{tang2024textsquare,lu2025bounding}, we propose the Context Excitation (CE) module to address the limitations of local feature representations in CNNs by capturing global contextual dependencies and performing adaptive channel recalibration.

Given an input feature map $\mathbf{F} \in \mathbb{R}^{C \times H \times W}$, the CE module first generates a spatial attention map through $1\times1$ convolution and softmax normalization:
\begin{equation}
\mathbf{F}_K = \text{Conv}_{1\times1}(\mathbf{F})
\label{eq:ce_key}
\end{equation}
\begin{equation}
\mathbf{F}_S = \text{softmax}(\mathbf{F}_K) \in \mathbb{R}^{1 \times H \times W}
\label{eq:ce_softmax}
\end{equation}
where the softmax operation normalizes the spatial responses to the range $[0,1]$, generating a probability distribution that highlights the most informative spatial locations.

The feature map $\mathbf{F}$ and the attention map $\mathbf{F}_S$ are then flattened and multiplied to aggregate spatially-weighted channel information:
\begin{equation}
\mathbf{F}_m = \text{matmul}(\text{flatten}(\mathbf{F}), \text{flatten}(\mathbf{F}_S)^T) \in \mathbb{R}^{C \times 1}
\label{eq:ce_matmul}
\end{equation}

The aggregated features undergo a bottleneck transformation with channel reduction and expansion, followed by sigmoid activation to produce channel-wise recalibration weights:
\begin{equation}
\mathbf{F}_T = \sigma(\text{Conv}_{1\times1}(\text{ReLU}(\text{Conv}_{1\times1}(\mathbf{F}_m)))) \in \mathbb{R}^{C \times 1 \times 1}
\label{eq:ce_recalib}
\end{equation}
where $\sigma(\cdot)$ denotes the sigmoid activation function.

The final output is obtained by element-wise multiplication of the recalibration weights with the original feature map:
\begin{equation}
\mathbf{F}_{\text{out}} = \mathbf{F}_T \cdot \mathbf{F}
\label{eq:ce_output}
\end{equation}

This mechanism enables the CE module to capture spatial inter-dependencies through the softmax-normalized attention map while performing channel-wise feature recalibration through the bottleneck transformation, effectively combining the strengths of spatial and channel attention in a computationally efficient manner. The CE module is positioned after the final MRDC layer (Mlayer4) to operate on the most semantically rich features, where contextual information is most beneficial for classification.

\subsection{MRDC Pooling Residual Block}

To integrate the MRDC module with the residual learning framework, we introduce the MRDC Pooling Residual Block (MRDC-block). Building upon the Average Pooling Residual Block (AP-block) that decouples the downsampling operation from convolution through average pooling on the skip connection, the MRDC-block replaces the standard convolution in the main path with the MRDC module:
\begin{equation}
H(\mathbf{x}) = M(f(\mathbf{x})) + g(\mathbf{x})
\label{eq:mrdc_block}
\end{equation}
where $f(\mathbf{x})$ represents the initial convolution and activation in the residual block, $M(\cdot)$ denotes the MRDC module, and $g(\mathbf{x})$ represents the skip connection with average pooling and $1\times1$ convolution for dimension matching. This design preserves the gradient flow properties of residual connections while leveraging the multi-scale feature extraction capabilities of MRDC.

\subsection{Overall Architecture}

The complete RDCNet architecture consists of five sequential stages:

\begin{enumerate}[leftmargin=*]
\item \textbf{Preprocessing:} Input images undergo data augmentation including random horizontal flipping, padding with random cropping, and random erasing~\cite{zhong2020random} to enhance sample diversity and reduce overfitting.

\item \textbf{Initial feature extraction:} Adaptive convolution based on input resolution (Section~3.1).

\item \textbf{Deep feature extraction:} Four layers of MRDC residual blocks (Mlayer1--Mlayer4) with progressively increasing channel depth, extracting hierarchical multi-scale features.

\item \textbf{Global feature integration:} The CE module processes the output of Mlayer4 to perform context-aware feature recalibration, followed by global average pooling to produce a fixed-size feature vector.

\item \textbf{Classification:} A fully connected layer maps the feature vector to class probabilities.
\end{enumerate}

\section{Experiments}

\subsection{Experimental Setup}

\textbf{Datasets.} We evaluate RDCNet on five widely-used benchmark datasets covering diverse visual recognition scenarios. (1)~\textbf{CIFAR-10}~contains 60,000 $32\times32$ color images across 10 natural image categories, split into 50,000 training and 10,000 test images. (2)~\textbf{CIFAR-100}~follows the same structure but with 100 fine-grained categories. (3)~\textbf{SVHN} (Street View House Numbers) contains over 99,000 $32\times32$ digit images from real-world street scenes, with 73,257 training and 26,032 test images. (4)~\textbf{Imagenette} is a curated 10-class subset of ImageNet with 9,469 training and 3,925 test images at $224\times224$ resolution. (5)~\textbf{Imagewoof} is a more challenging 10-class ImageNet subset focusing on dog breeds, with 9,025 training and 3,929 test images.

\textbf{Implementation details.} All experiments are conducted using PyTorch 2.3.0 on an NVIDIA RTX 3090 GPU with CUDA 11.8. We employ SGD optimization with momentum 0.9, weight decay $5\times10^{-4}$, and label smoothing of 0.1. The initial learning rate is set to 0.1 with cosine annealing~\cite{loshchilov2017sgdr} scheduling over 200 epochs. Batch sizes are 128 for $32\times32$ inputs and 64 for $224\times224$ inputs. Data augmentation includes random horizontal flipping, padding with random cropping, and random erasing~\cite{zhong2020random}.

\textbf{Evaluation metric.} Classification accuracy (top-1) on the test set is used as the primary evaluation metric across all experiments.

\subsection{Analysis of MRDC Module Parameters}

\subsubsection{Impact of Masking Strategy}

To investigate the contribution of different masking components within the MRDC module, we conduct a systematic ablation study comparing four masking strategies: no masking (null\_mask), kernel masking only (k\_mask), channel masking only (c\_mask), and combined channel-kernel masking (c\_k\_mask). The results in Table~\ref{tab:mask_strategy} demonstrate that the combined masking strategy consistently achieves the best performance across all five datasets.

\begin{table}[t]
\centering
\caption{Impact of different masking strategies on MRDC module performance (classification accuracy \%).}
\label{tab:mask_strategy}
\small
\begin{tabular}{lccccc}
\toprule
Strategy & C-10 & C-100 & SVHN & Inette & Iwoof \\
\midrule
null\_mask & 96.08 & 79.98 & 96.92 & 90.93 & 83.84 \\
k\_mask   & 96.24 & 80.41 & 97.23 & 91.72 & 84.93 \\
c\_mask   & 96.16 & 80.76 & 97.19 & 91.11 & 85.03 \\
c\_k\_mask & \textbf{96.38} & \textbf{81.16} & \textbf{97.41} & \textbf{92.58} & \textbf{85.14} \\
\bottomrule
\end{tabular}
\end{table}

The superiority of the combined masking strategy can be attributed to the complementary effects of channel and kernel masking. Channel masking encourages the network to learn robust representations across feature channels, preventing over-reliance on any single channel. Kernel masking, on the other hand, disrupts the regular spatial sampling pattern of dilated convolutions, effectively mitigating gridding artifacts. When applied jointly, these two masking mechanisms create a rich space of feature interaction patterns that enhance both the diversity and robustness of learned representations.

\subsubsection{Impact of Scaling Factor}

The scaling factor $\alpha$ in Equation~\ref{eq:mrdc_output} controls the relative contribution of the multi-branch features versus the residual shortcut. We systematically evaluate $\alpha$ values ranging from 0.1 to 1.0 across all five datasets, with results presented in Table~\ref{tab:scale}.

\begin{table}[t]
\centering
\caption{Impact of scaling factor $\alpha$ on MRDC module performance (classification accuracy \%).}
\label{tab:scale}
\small
\begin{tabular}{lccccc}
\toprule
$\alpha$ & C-10 & C-100 & SVHN & Inette & Iwoof \\
\midrule
0.1 & 96.09 & 80.30 & 97.23 & 92.25 & 84.35 \\
0.2 & 96.16 & 80.78 & 97.25 & 92.15 & 84.19 \\
0.3 & 96.22 & 81.09 & 97.26 & 91.90 & 83.76 \\
0.4 & 96.27 & 80.95 & \textbf{97.41} & 92.23 & 84.91 \\
0.5 & \textbf{96.38} & \textbf{81.16} & 97.39 & \textbf{92.58} & 85.09 \\
0.6 & 95.89 & 80.84 & 97.34 & 92.48 & \textbf{85.14} \\
0.7 & 95.78 & 80.68 & 97.38 & 92.20 & 84.58 \\
0.8 & 95.86 & 80.31 & 97.29 & 92.03 & 84.68 \\
0.9 & 95.72 & 79.70 & 97.18 & 91.54 & 83.89 \\
1.0 & 95.75 & 80.01 & 97.13 & 91.46 & 84.34 \\
\bottomrule
\end{tabular}
\end{table}

The results reveal that the optimal $\alpha$ value centers around 0.5, indicating that balanced integration of multi-branch features with the residual input yields the best classification performance. When $\alpha$ is too small, the network over-relies on the original features, which are insufficiently rich to capture complex contextual patterns. Conversely, when $\alpha$ is too large, the network becomes overly dependent on the complex multi-branch features, increasing sensitivity to noise and overfitting risk. The moderate value of $\alpha \approx 0.5$ provides an effective equilibrium, allowing the network to leverage both low-level residual features and high-level multi-scale representations.

\subsection{Placement and Quantity of CE Module}

The positioning of the CE module within the network architecture significantly affects its effectiveness. We design eight configuration variants (A--H) with CE modules placed at different layers (Mlayer1--Mlayer4) and in different combinations. The results in Table~\ref{tab:ce_position} indicate that configuration E, which places a single CE module after Mlayer4, consistently achieves the highest accuracy across all datasets.

\begin{table}[t]
\centering
\caption{Impact of CE module placement on classification accuracy (\%). Configuration E places CE after Mlayer4 only.}
\label{tab:ce_position}
\small
\begin{tabular}{lccccc}
\toprule
Config & C-10 & C-100 & SVHN & Inette & Iwoof \\
\midrule
A (none) & 96.38 & 81.16 & 97.41 & 92.58 & 85.14 \\
B & 96.58 & 80.15 & 97.64 & 92.18 & 84.39 \\
C & 96.47 & 80.87 & 97.71 & 91.26 & 84.65 \\
D & 96.26 & 80.48 & 97.73 & 92.38 & 85.44 \\
E & \textbf{96.71} & \textbf{81.75} & \textbf{97.78} & \textbf{93.48} & \textbf{85.65} \\
F & 96.18 & 81.05 & 97.74 & 91.89 & 85.46 \\
G & 96.30 & 80.12 & 97.71 & 92.76 & 85.11 \\
H & 96.40 & 79.81 & 97.57 & 92.25 & 84.98 \\
\bottomrule
\end{tabular}
\end{table}

This finding is consistent with the intuition that deeper layers contain more abstract and semantically rich features. Feature maps at Mlayer4 encode high-level semantic information about object categories, making them the most suitable candidates for context-aware recalibration. At shallower layers, features are predominantly low-level (edges, textures) and lack sufficient semantic content to benefit from contextual attention. Moreover, applying CE at deeper layers leverages the smaller spatial dimensions and larger channel counts of these features, providing the CE module with richer contextual information for effective channel recalibration.

\subsection{Comparison of CE with Alternative Modules}

To further validate the superiority of the proposed CE module, we compare it against four established attention and context modules: SE~\cite{hu2018squeeze}, Non-Local Block (NLB)~\cite{wang2018non}, Criss-Cross Attention (CCA)~\cite{huang2019ccnet}, and Global Context Block (GCB)~\cite{cao2019gcnet}. Each module is placed in the same position (after Mlayer4) within RDCNet. As shown in Table~\ref{tab:ce_compare}, the CE module achieves the highest accuracy on all five datasets.

\begin{table}[t]
\centering
\caption{Comparison of different attention modules within RDCNet (classification accuracy \%).}
\label{tab:ce_compare}
\small
\begin{tabular}{lccccc}
\toprule
Module & C-10 & C-100 & SVHN & Inette & Iwoof \\
\midrule
+ SE  & 96.55 & 81.28 & 97.35 & 93.09 & 85.14 \\
+ NLB & 96.13 & 81.10 & 97.58 & 91.80 & 85.39 \\
+ CCA & 96.00 & 80.15 & 97.26 & 93.07 & 85.11 \\
+ GCB & 96.48 & 80.59 & 97.66 & 92.97 & 84.58 \\
+ CE  & \textbf{96.71} & \textbf{81.75} & \textbf{97.78} & \textbf{93.48} & \textbf{85.65} \\
\bottomrule
\end{tabular}
\end{table}

The CE module outperforms SE by capturing spatial dependencies in addition to channel relationships. Compared to NLB, CE achieves superior performance with significantly lower computational cost by avoiding dense pairwise computations. The advantage over CCA stems from CE's ability to capture holistic global context rather than being limited to criss-cross paths. Finally, CE surpasses GCB by dynamically balancing global and local information through its excitation mechanism, avoiding the over-globalization that can afflict purely global context approaches.

\subsection{Impact of Network Depth and Learning Rate}

We investigate the effect of network depth by comparing RDCNet configurations with 18, 34, 50, and 101 layers on CIFAR-100. As shown in Table~\ref{tab:depth}, RDCNet34 achieves the best performance with 81.75\% accuracy, outperforming RDCNet18 by 2.36\%, RDCNet50 by 1.64\%, and RDCNet101 by 2.32\%.

\begin{table}[t]
\centering
\caption{Comparison of different network depths on CIFAR-100.}
\label{tab:depth}
\small
\begin{tabular}{lccc}
\toprule
Network & Accuracy (\%) & F1-Score & X-Entropy \\
\midrule
RDCNet18  & 79.39 & 0.7937 & 0.231 \\
RDCNet34  & \textbf{81.75} & \textbf{0.8172} & \textbf{0.151} \\
RDCNet50  & 80.11 & 0.8009 & 0.234 \\
RDCNet101 & 79.43 & 0.7940 & 0.230 \\
\bottomrule
\end{tabular}
\end{table}

The 34-layer configuration represents the optimal balance between model capacity and generalization on CIFAR-100. Shallower networks (18 layers) lack sufficient representational capacity for the 100-class classification task, while deeper networks (50 and 101 layers) suffer from increased overfitting risk due to their larger parameter counts, which is exacerbated by the relatively small size of the CIFAR-100 training set.

We also study the impact of the initial learning rate under cosine annealing scheduling, evaluating five values (0.1, 0.05, 0.01, 0.005, 0.001) on CIFAR-10, CIFAR-100, and SVHN. An initial learning rate of 0.1 consistently yields the best performance across all three datasets, enabling rapid convergence in early training phases while the cosine annealing schedule ensures fine-grained parameter optimization in later stages.

\subsection{Comparison with State-of-the-Art Methods}

We compare RDCNet against 14 representative classification methods spanning CNNs, transformers, and hybrid architectures. The comprehensive results in Table~\ref{tab:sota} demonstrate that RDCNet achieves the highest classification accuracy on all five benchmark datasets.

\begin{table*}[t]
\centering
\caption{Classification accuracy (\%) comparison with state-of-the-art methods on five benchmark datasets. Best results are in \textbf{bold}. ``--'' indicates results not reported in the original paper.}
\label{tab:sota}
\small
\begin{tabular}{lccccc}
\toprule
Method & CIFAR-10 & CIFAR-100 & SVHN & Imagenette & Imagewoof \\
\midrule
ResNet-34~\cite{he2016deep} & 88.87 & 71.49 & 95.51 & 87.72 & 78.04 \\
HO-ResNet~\cite{luo2022rethinking} & 96.32 & 77.12 & 95.69 & 87.23 & 79.74 \\
WRN-28-10~\cite{zagoruyko2017wide} & 95.87 & 80.50 & 96.58 & 88.34 & 78.71 \\
CAPR-DenseNet & 94.24 & 78.85 & 94.97 & 87.72 & 80.78 \\
Multi-ResNet~\cite{abdi2017multi} & 94.56 & 78.69 & 94.58 & 87.71 & 81.26 \\
EfficientNet~\cite{tan2020efficientnet} & 94.02 & 75.98 & 93.34 & 88.06 & 77.95 \\
GhostNet~\cite{han2020ghostnet} & 94.92 & 77.17 & 93.86 & 87.83 & 78.35 \\
Couplformer~\cite{lan2021couplformer} & 93.54 & 73.92 & 94.26 & 87.91 & 77.89 \\
FAVOR+~\cite{choromanski2020rethinking} & 91.42 & 72.56 & 93.21 & 88.16 & 77.57 \\
MMA-CCT~\cite{konstantinidis2023multi} & 94.74 & 77.51 & 94.29 & -- & -- \\
ATONet~\cite{wu2024auto} & 94.51 & 78.56 & 95.21 & 86.68 & 80.21 \\
QKFormer~\cite{zhou2024qkformer} & 96.18 & 80.27 & 97.13 & 88.42 & 81.67 \\
DAMSNet~\cite{jiang2023double} & 96.51 & 80.50 & 97.60 & -- & -- \\
SSCNet~\cite{jiang2024sparse} & 96.69 & 80.63 & 97.43 & 88.75 & 82.09 \\
\midrule
\textbf{RDCNet (Ours)} & \textbf{96.71} & \textbf{81.75} & \textbf{97.78} & \textbf{93.48} & \textbf{85.65} \\
\bottomrule
\end{tabular}
\end{table*}

Notably, RDCNet demonstrates particularly significant improvements on the more challenging Imagenette and Imagewoof datasets, where it surpasses the second-best method by 4.73\% and 3.56\% respectively. These larger margins on higher-resolution datasets with more complex backgrounds validate the effectiveness of our multi-scale feature extraction and context-aware attention mechanisms in handling real-world visual complexity. On the smaller-scale CIFAR and SVHN datasets, RDCNet still achieves consistent improvements, confirming its generalizability across different data scales and visual domains.

Compared with the baseline ResNet-34, RDCNet achieves remarkable improvements of 7.84\%, 10.26\%, 2.27\%, 5.76\%, and 7.61\% on the five datasets respectively, underscoring the substantial enhancement provided by the proposed MRDC and CE modules. The consistent superiority over recent transformer-based methods (Couplformer, FAVOR+, QKFormer) demonstrates that well-designed convolutional architectures with attention mechanisms can remain highly competitive against the transformer paradigm for image classification.

\subsection{Ablation Studies}

To systematically assess the contribution of each component in RDCNet, we conduct comprehensive ablation experiments. Four ablation configurations are evaluated: (1)~\textbf{Net1}: RDCNet without the CE module; (2)~\textbf{Net2}: RDCNet with MRDC replaced by standard $3\times3$ convolutions; (3)~\textbf{Net3}: Both CE and MRDC removed; (4)~\textbf{Net4}: RDCNet without the modified initial convolution (retaining $7\times7$ kernel and max pooling).

\begin{table}[t]
\centering
\caption{Ablation study results (classification accuracy \%).}
\label{tab:ablation}
\small
\begin{tabular}{lccccc}
\toprule
Model & C-10 & C-100 & SVHN & Inette & Iwoof \\
\midrule
RDCNet & \textbf{96.71} & \textbf{81.75} & \textbf{97.78} & \textbf{93.48} & \textbf{85.65} \\
Net1 (w/o CE) & 96.38 & 81.16 & 97.41 & 92.58 & 85.14 \\
Net2 (w/o MRDC) & 95.46 & 79.78 & 97.03 & 90.75 & 81.06 \\
Net3 (w/o both) & 95.23 & 78.35 & 96.65 & -- & -- \\
Net4 (7$\times$7 init) & 88.87 & 71.49 & 95.51 & 87.72 & 78.04 \\
\bottomrule
\end{tabular}
\end{table}

The ablation results in Table~\ref{tab:ablation} yield several important insights. First, removing the CE module (Net1) results in accuracy drops of 0.33\%, 0.59\%, 0.37\%, 0.90\%, and 0.51\% across the five datasets, confirming that contextual feature recalibration contributes meaningfully to classification performance. Second, removing the MRDC module (Net2) causes more substantial degradation of 1.25\%, 1.97\%, 0.75\%, 2.73\%, and 4.59\%, demonstrating that multi-scale feature extraction with random masking is the primary driver of RDCNet's improved performance. Third, removing both modules (Net3) leads to the largest accuracy drops, indicating that MRDC and CE provide complementary and synergistic benefits. Finally, the dramatic performance gap between RDCNet and Net4 on small-resolution datasets validates the importance of the adaptive initial convolution design.

\section{Conclusion}

We have presented RDCNet, a novel image classification architecture that addresses the long-standing challenges of fine-grained feature extraction, background noise suppression, and overfitting through three synergistic innovations built upon the ResNet-34 backbone. The Multi-Branch Random Dilated Convolution (MRDC) module employs parallel branches with diverse dilation rates and a stochastic masking mechanism to capture multi-scale features while mitigating gridding artifacts and providing implicit regularization, enhanced by the Fine-Grained Feature Enhancement (FGFE) branch that bridges global contextual information with local feature representations to amplify subtle discriminative patterns. The Context Excitation (CE) module further strengthens the network by combining softmax-based spatial attention with channel recalibration to dynamically emphasize task-relevant features while suppressing background interference, achieving a favorable balance between computational efficiency and representational capacity. Comprehensive experiments on five diverse benchmark datasets demonstrate that RDCNet consistently achieves state-of-the-art classification performance, with improvements of up to 10.26\% over the ResNet-34 baseline and significant margins over 14 competing methods including recent transformer-based architectures, while ablation studies confirm the complementary contributions of each proposed module. Despite these encouraging results, computational overhead from the multi-branch architecture and the balance between local and global feature capture in heavily cluttered scenes remain areas for future improvement, alongside extending RDCNet to other visual recognition tasks such as object detection~\cite{tang2022few,liu2018receptive}, scene text recognition~\cite{tang2022optimal,zhao2023multi,liu2023spts}, and document understanding~\cite{feng2023docpedia,tang2024textsquare,feng2025dolphin,wang2025wilddoc,tang2025mtvqa,zhao2024harmonizing,zhao2024tabpedia,lu2025bounding,tang2023character,tang2022you,fei2025advancing,wang2025vision,shan2024mctbench,wang2024pargo,sun2024attentive,feng2023unidoc,feng2026dolphin_v2,zhu2026textpecker}.

{\small
\bibliographystyle{ieee_fullname}
\bibliography{references}

@inproceedings{lecun1998gradient,
  title={Gradient-based learning applied to document recognition},
  author={LeCun, Yann and Bottou, L{\'e}on and Bengio, Yoshua and Haffner, Patrick},
  journal={Proceedings of the IEEE},
  volume={86},
  number={11},
  pages={2278--2324},
  year={1998},
  publisher={IEEE}
}

@article{krizhevsky2017imagenet,
  title={ImageNet classification with deep convolutional neural networks},
  author={Krizhevsky, Alex and Sutskever, Ilya and Hinton, Geoffrey E},
  journal={Communications of the ACM},
  volume={60},
  number={6},
  pages={84--90},
  year={2017},
  publisher={ACM}
}

@article{simonyan2015very,
  title={Very deep convolutional networks for large-scale image recognition},
  author={Simonyan, Karen and Zisserman, Andrew},
  journal={arXiv preprint arXiv:1409.1556},
  year={2015}
}

@inproceedings{szegedy2015going,
  title={Going deeper with convolutions},
  author={Szegedy, Christian and Liu, Wei and Jia, Yangqing and Sermanet, Pierre and Reed, Scott and Anguelov, Dragomir and Erhan, Dumitru and Vanhoucke, Vincent and Rabinovich, Andrew},
  booktitle={Proceedings of the IEEE Conference on Computer Vision and Pattern Recognition},
  pages={1--9},
  year={2015}
}

@inproceedings{he2016deep,
  title={Deep residual learning for image recognition},
  author={He, Kaiming and Zhang, Xiangyu and Ren, Shaoqing and Sun, Jian},
  booktitle={Proceedings of the IEEE Conference on Computer Vision and Pattern Recognition},
  pages={770--778},
  year={2016}
}

@article{zagoruyko2017wide,
  title={Wide residual networks},
  author={Zagoruyko, Sergey and Komodakis, Nikos},
  journal={arXiv preprint arXiv:1605.07146},
  year={2017}
}

@inproceedings{huang2017densely,
  title={Densely connected convolutional networks},
  author={Huang, Gao and Liu, Zhuang and Van Der Maaten, Laurens and Weinberger, Kilian Q},
  booktitle={Proceedings of the IEEE Conference on Computer Vision and Pattern Recognition},
  pages={2261--2269},
  year={2017}
}

@article{howard2017mobilenets,
  title={MobileNets: Efficient convolutional neural networks for mobile vision applications},
  author={Howard, Andrew G and Zhu, Menglong and Chen, Bo and Kalenichenko, Dmitry and Wang, Weijun and Weyand, Tobias and Andreetto, Marco and Adam, Hartwig},
  journal={arXiv preprint arXiv:1704.04861},
  year={2017}
}

@article{wang2022dual,
  title={Dual-path processing network for high-resolution salient object detection},
  author={Wang, Jian and Yang, Qiping and Yang, Shiqiang and Chai, Xiuli and Zhang, Wenjie},
  journal={Applied Intelligence},
  volume={52},
  number={10},
  pages={12034--12048},
  year={2022},
  publisher={Springer}
}

@article{abdi2017multi,
  title={Multi-residual networks: Improving the speed and accuracy of residual networks},
  author={Abdi, Masoud and Nahavandi, Saeid},
  journal={arXiv preprint arXiv:1609.05672},
  year={2017}
}

@article{luo2022rethinking,
  title={Rethinking ResNets: Improved stacking strategies with high-order schemes for image classification},
  author={Luo, Zhibo and Sun, Zhitao and Zhou, Weilun and Wu, Zhengzhong and Kamata, Sei-ichiro},
  journal={Complex and Intelligent Systems},
  volume={8},
  number={4},
  pages={3395--3407},
  year={2022},
  publisher={Springer}
}

@article{chen2018deeplab,
  title={DeepLab: Semantic image segmentation with deep convolutional nets, atrous convolution, and fully connected CRFs},
  author={Chen, Liang-Chieh and Papandreou, George and Kokkinos, Iasonas and Murphy, Kevin and Yuille, Alan L},
  journal={IEEE Transactions on Pattern Analysis and Machine Intelligence},
  volume={40},
  number={4},
  pages={834--848},
  year={2018},
  publisher={IEEE}
}

@inproceedings{liu2018receptive,
  title={Receptive field block net for accurate and fast object detection},
  author={Liu, Songtao and Huang, Di and Wang, Yunhong},
  booktitle={Proceedings of the European Conference on Computer Vision},
  pages={404--419},
  year={2018},
  publisher={Springer}
}

@inproceedings{chen2024frequency,
  title={Frequency-adaptive dilated convolution for semantic segmentation},
  author={Chen, Linwei and Gu, Lin and Zheng, Dezhi and Fu, Ying},
  booktitle={Proceedings of the IEEE/CVF Conference on Computer Vision and Pattern Recognition},
  pages={3414--3425},
  year={2024}
}

@inproceedings{jiang2025when,
  title={When fast fourier transform meets transformer for image restoration},
  author={Jiang, Xueyang and Zhang, Xiaohan and Gao, Nan and Deng, Yue},
  booktitle={Proceedings of the European Conference on Computer Vision},
  pages={381--402},
  year={2025},
  publisher={Springer}
}

@inproceedings{vaswani2017attention,
  title={Attention is all you need},
  author={Vaswani, Ashish and Shazeer, Noam and Parmar, Niki and Uszkoreit, Jakob and Jones, Llion and Gomez, Aidan N and Kaiser, {\L}ukasz and Polosukhin, Illia},
  booktitle={Advances in Neural Information Processing Systems},
  pages={6000--6010},
  year={2017}
}

@inproceedings{wang2018non,
  title={Non-local neural networks},
  author={Wang, Xiaolong and Girshick, Ross and Gupta, Abhinav and He, Kaiming},
  booktitle={Proceedings of the IEEE/CVF Conference on Computer Vision and Pattern Recognition},
  year={2018}
}

@inproceedings{huang2019ccnet,
  title={CCNet: Criss-cross attention for semantic segmentation},
  author={Huang, Zilong and Wang, Xinggang and Huang, Lichao and Huang, Chang and Wei, Yunchao and Liu, Wenyu},
  booktitle={Proceedings of the IEEE/CVF International Conference on Computer Vision},
  pages={603--612},
  year={2019}
}

@inproceedings{zhu2019asymmetric,
  title={Asymmetric non-local neural networks for semantic segmentation},
  author={Zhu, Zhen and Xu, Mengde and Bai, Song and Huang, Tengteng and Bai, Xiang},
  booktitle={Proceedings of the IEEE/CVF International Conference on Computer Vision},
  pages={593--602},
  year={2019}
}

@inproceedings{hu2018squeeze,
  title={Squeeze-and-excitation networks},
  author={Hu, Jie and Shen, Li and Sun, Gang},
  booktitle={Proceedings of the IEEE/CVF Conference on Computer Vision and Pattern Recognition},
  pages={7132--7141},
  year={2018}
}

@inproceedings{cao2019gcnet,
  title={GCNet: Non-local networks meet squeeze-excitation networks and beyond},
  author={Cao, Yue and Xu, Jiarui and Lin, Stephen and Wei, Fangyun and Hu, Han},
  booktitle={Proceedings of the IEEE/CVF International Conference on Computer Vision Workshop},
  pages={1971--1980},
  year={2019}
}

@article{zhong2020random,
  title={Random erasing data augmentation},
  author={Zhong, Zhun and Zheng, Liang and Kang, Guoliang and Li, Shaozi and Yang, Yi},
  journal={Proceedings of the AAAI Conference on Artificial Intelligence},
  volume={34},
  number={7},
  pages={13001--13008},
  year={2020}
}

@article{tan2020efficientnet,
  title={EfficientNet: Rethinking model scaling for convolutional neural networks},
  author={Tan, Mingxing and Le, Quoc V},
  journal={arXiv preprint arXiv:1905.11946},
  year={2020}
}

@inproceedings{han2020ghostnet,
  title={GhostNet: More features from cheap operations},
  author={Han, Kai and Wang, Yunhe and Tian, Qi and Guo, Jianyuan and Xu, Chunjing and Xu, Chang},
  booktitle={Proceedings of the IEEE/CVF Conference on Computer Vision and Pattern Recognition},
  pages={1577--1586},
  year={2020}
}

@inproceedings{lan2021couplformer,
  title={Couplformer: Rethinking vision transformer with coupling attention},
  author={Lan, Hao and Wang, Xiaohu and Shen, Hao and Liang, Pengda and Wei, Xian},
  booktitle={Proceedings of the IEEE/CVF Winter Conference on Applications of Computer Vision},
  year={2023}
}

@article{choromanski2020rethinking,
  title={Rethinking attention with performers},
  author={Choromanski, Krzysztof and Likhosherstov, Valerii and Dohan, David and Song, Xingyou and Gane, Andreea and Sarlos, Tamas and Hawkins, Peter and Davis, Jared and Mohiuddin, Afroz and Kaiser, Lukasz and others},
  journal={arXiv preprint arXiv:2009.14794},
  year={2020}
}

@article{konstantinidis2023multi,
  title={Multi-manifold attention for vision transformers},
  author={Konstantinidis, Dimitrios and Papastratis, Ilias and Dimitropoulos, Kosmas and Daras, Petros},
  journal={IEEE Access},
  volume={11},
  pages={123433--123444},
  year={2023},
  publisher={IEEE}
}

@inproceedings{wu2024auto,
  title={Auto-train-once: Controller network guided automatic network pruning from scratch},
  author={Wu, Xidong and Gao, Shangqian and Zhang, Zeyu and Li, Zhenzhen and Bao, Runxue and Zhang, Yanfu and Wang, Xiaoqian and Huang, Heng},
  booktitle={Proceedings of the IEEE/CVF Conference on Computer Vision and Pattern Recognition},
  pages={16163--16173},
  year={2024}
}

@article{zhou2024qkformer,
  title={QKFormer: Hierarchical spiking transformer using Q-K attention},
  author={Zhou, Chenlin and Zhang, Han and Zhou, Zhaokun and Yu, Liutao and Huang, Liwei and Fan, Xiaopeng and Yuan, Li and Ma, Zhengyu and Zhou, Huihui and Tian, Yonghong},
  journal={arXiv preprint arXiv:2403.16552},
  year={2024}
}

@article{jiang2024sparse,
  title={Sparse feature image classification network with spatial position correction},
  author={Jiang, Wentao and Chen, Chen and Zhang, Shengchong},
  journal={Opto-Electronic Engineering},
  volume={51},
  number={5},
  pages={240050},
  year={2024}
}

@article{jiang2023double,
  title={Double-branch multi-attention mechanism based sharpness-aware classification network},
  author={Jiang, Wentao and Zhao, Linlin and Tu, Chao},
  journal={Pattern Recognition and Artificial Intelligence},
  volume={36},
  number={3},
  pages={252--267},
  year={2023}
}

@inproceedings{dosovitskiy2021image,
  title={An image is worth 16x16 words: Transformers for image recognition at scale},
  author={Dosovitskiy, Alexey and Beyer, Lucas and Kolesnikov, Alexander and Weissenborn, Dirk and Zhai, Xiaohua and Unterthiner, Thomas and Dehghani, Mostafa and Minderer, Matthias and Heigold, Georg and Gelly, Sylvain and others},
  booktitle={International Conference on Learning Representations},
  year={2021}
}

@inproceedings{liu2021swin,
  title={Swin transformer: Hierarchical vision transformer using shifted windows},
  author={Liu, Ze and Lin, Yutong and Cao, Yue and Hu, Han and Wei, Yixuan and Zhang, Zheng and Lin, Stephen and Guo, Baining},
  booktitle={Proceedings of the IEEE/CVF International Conference on Computer Vision},
  pages={10012--10022},
  year={2021}
}

@inproceedings{woo2018cbam,
  title={CBAM: Convolutional block attention module},
  author={Woo, Sanghyun and Park, Jongchan and Lee, Joon-Young and Kweon, In So},
  booktitle={Proceedings of the European Conference on Computer Vision},
  pages={3--19},
  year={2018}
}

@article{yu2016multi,
  title={Multi-scale context aggregation by dilated convolutions},
  author={Yu, Fisher and Koltun, Vladlen},
  journal={arXiv preprint arXiv:1511.07122},
  year={2016}
}

@inproceedings{chen2017rethinking,
  title={Rethinking atrous convolution for semantic image segmentation},
  author={Chen, Liang-Chieh and Papandreou, George and Schroff, Florian and Adam, Hartwig},
  journal={arXiv preprint arXiv:1706.05587},
  year={2017}
}

@article{srivastava2014dropout,
  title={Dropout: A simple way to prevent neural networks from overfitting},
  author={Srivastava, Nitish and Hinton, Geoffrey and Krizhevsky, Alex and Sutskever, Ilya and Salakhutdinov, Ruslan},
  journal={The Journal of Machine Learning Research},
  volume={15},
  number={1},
  pages={1929--1958},
  year={2014}
}

@inproceedings{dai2017deformable,
  title={Deformable convolutional networks},
  author={Dai, Jifeng and Qi, Haozhi and Xiong, Yuwen and Li, Yi and Zhang, Guodong and Hu, Han and Wei, Yichen},
  booktitle={Proceedings of the IEEE International Conference on Computer Vision},
  pages={764--773},
  year={2017}
}

@inproceedings{sandler2018mobilenetv2,
  title={MobileNetV2: Inverted residuals and linear bottlenecks},
  author={Sandler, Mark and Howard, Andrew and Zhu, Menglong and Zhmoginov, Andrey and Chen, Liang-Chieh},
  booktitle={Proceedings of the IEEE Conference on Computer Vision and Pattern Recognition},
  pages={4510--4520},
  year={2018}
}

@inproceedings{ma2018shufflenet,
  title={ShuffleNet V2: Practical guidelines for efficient CNN architecture design},
  author={Ma, Ningning and Zhang, Xiangyu and Zheng, Hai-Tao and Sun, Jian},
  booktitle={Proceedings of the European Conference on Computer Vision},
  pages={116--131},
  year={2018}
}

@inproceedings{loshchilov2017sgdr,
  title={SGDR: Stochastic gradient descent with warm restarts},
  author={Loshchilov, Ilya and Hutter, Frank},
  booktitle={International Conference on Learning Representations},
  year={2017}
}

@inproceedings{devries2017improved,
  title={Improved regularization of convolutional neural networks with cutout},
  author={DeVries, Terrell and Taylor, Graham W},
  journal={arXiv preprint arXiv:1708.04552},
  year={2017}
}

@inproceedings{yun2019cutmix,
  title={CutMix: Regularization strategy to train strong classifiers with localizable features},
  author={Yun, Sangdoo and Han, Dongyoon and Oh, Seong Joon and Chun, Sanghyuk and Choe, Junsuk and Yoo, Youngjoon},
  booktitle={Proceedings of the IEEE/CVF International Conference on Computer Vision},
  pages={6023--6032},
  year={2019}
}

@inproceedings{zhang2018mixup,
  title={mixup: Beyond empirical risk minimization},
  author={Zhang, Hongyi and Cisse, Moustapha and Dauphin, Yann N and Lopez-Paz, David},
  booktitle={International Conference on Learning Representations},
  year={2018}
}

@inproceedings{tang2022few,
  title={Few could be better than all: Feature sampling and grouping for scene text detection},
  author={Tang, Jingqun and Zhang, Wenqing and Liu, Hongye and Yang, Min-Kuang and Jiang, Bo and Hu, Guanglong and Bai, Xiang},
  booktitle={Proceedings of the IEEE/CVF Conference on Computer Vision and Pattern Recognition},
  pages={4563--4572},
  year={2022}
}

@inproceedings{tang2022optimal,
  title={Optimal boxes: Boosting end-to-end scene text recognition by adjusting annotated bounding boxes via reinforcement learning},
  author={Tang, Jingqun and Qian, Wenming and Song, Luchuan and Dong, Xiena and Li, Lan and Bai, Xiang},
  booktitle={Proceedings of the European Conference on Computer Vision},
  pages={233--248},
  year={2022},
  publisher={Springer}
}

@inproceedings{tang2022you,
  title={You can even annotate text with voice: Transcription-only-supervised text spotting},
  author={Tang, Jingqun and Qiao, Shaohua and Cui, Benlei and Ma, Yuhang and Zhang, Sheng and Kanoulas, Dimitrios},
  booktitle={Proceedings of the 30th ACM International Conference on Multimedia},
  pages={4154--4163},
  year={2022}
}

@article{tang2024textsquare,
  title={TextSquare: Scaling up text-centric visual instruction tuning},
  author={Tang, Jingqun and Lin, Chunhui and Zhao, Zhen and Wei, Shuai and Wu, Binghong and Liu, Qi and He, Yue and Lu, Keren and Feng, Hao and Li, Yuliang and others},
  journal={arXiv preprint arXiv:2404.12803},
  year={2024}
}

@article{tang2025mtvqa,
  title={MTVQA: Benchmarking multilingual text-centric visual question answering},
  author={Tang, Jingqun and Liu, Qi and Ye, Yongjie and Lu, Jinghui and Wei, Shuai and Wang, Anlong and Lin, Chunhui and Feng, Hao and Zhao, Zhen and others},
  journal={Findings of the Association for Computational Linguistics: ACL 2025},
  pages={7748--7763},
  year={2025}
}

@article{feng2023docpedia,
  title={DocPedia: Unleashing the power of large multimodal model in the frequency domain for versatile document understanding},
  author={Feng, Hao and Liu, Qi and Liu, Hao and Tang, Jingqun and Zhou, Wengang and Li, Houqiang and Huang, Can},
  journal={Science China Information Sciences},
  year={2023}
}

@article{feng2023unidoc,
  title={UniDoc: A universal large multimodal model for simultaneous text detection, recognition, spotting and understanding},
  author={Feng, Hao and Wang, Zijian and Tang, Jingqun and Lu, Jinghui and Zhou, Wengang and Li, Houqiang and Huang, Can},
  journal={arXiv preprint arXiv:2308.11592},
  year={2023}
}

@inproceedings{zhao2023multi,
  title={Multi-modal in-context learning makes an ego-evolving scene text recognizer},
  author={Zhao, Zhen and Tang, Jingqun and Wu, Binghong and Lin, Chunhui and Liu, Hao and Zhang, Zitao and Tan, Xin and Huang, Can and Xie, Yuan},
  booktitle={Proceedings of the IEEE/CVF Conference on Computer Vision and Pattern Recognition},
  year={2023}
}

@inproceedings{zhao2024harmonizing,
  title={Harmonizing visual text comprehension and generation},
  author={Zhao, Zhen and Tang, Jingqun and Wu, Binghong and Lin, Chunhui and Wei, Shuai and Liu, Hao and Tan, Xin and Zhang, Zitao and Huang, Can and others},
  booktitle={Advances in Neural Information Processing Systems},
  year={2024}
}

@inproceedings{zhao2024tabpedia,
  title={TabPedia: Towards comprehensive visual table understanding with concept synergy},
  author={Zhao, Weichao and Feng, Hao and Liu, Qi and Tang, Jingqun and Wei, Shuai and Wu, Binghong and Liao, Lei and Ye, Yongjie and Liu, Hao and Zhou, Wengang and others},
  booktitle={Advances in Neural Information Processing Systems},
  year={2024}
}

@article{tang2023character,
  title={Character recognition competition for street view shop signs},
  author={Tang, Jingqun and Du, Wei and Wang, Bo and Zhou, Wengang and Mei, Songlin and Xue, Tian and Xu, Xin and Zhang, Hao},
  journal={National Science Review},
  volume={10},
  number={6},
  pages={nwad141},
  year={2023},
  publisher={Oxford University Press}
}

@article{liu2023spts,
  title={SPTS v2: Single-point scene text spotting},
  author={Liu, Yuliang and Zhang, Jiaxin and Peng, Dezhi and Huang, Mingxin and Wang, Xinyu and Tang, Jingqun and Huang, Can and Lin, Dahua and others},
  journal={IEEE Transactions on Pattern Analysis and Machine Intelligence},
  volume={45},
  number={12},
  year={2023},
  publisher={IEEE}
}

@inproceedings{feng2025dolphin,
  title={Dolphin: Document image parsing via heterogeneous anchor prompting},
  author={Feng, Hao and Wei, Shuai and Fei, Xingjian and Shi, Wenqiang and Han, Yuechen and Liao, Lei and Lu, Jinghui and Wu, Binghong and Liu, Qi and Lin, Chunhui and others},
  booktitle={Findings of the Association for Computational Linguistics: ACL 2025},
  pages={21919--21936},
  year={2025}
}

@article{lu2025bounding,
  title={A bounding box is worth one token -- interleaving layout and text in a large language model for document understanding},
  author={Lu, Jinghui and Yu, Haiyang and Wang, Yanjie and Ye, Yongjie and Tang, Jingqun and Yang, Ziwei and Wu, Binghong and Liu, Qi and Feng, Hao and Wang, Han and others},
  journal={Findings of the Association for Computational Linguistics: ACL 2025},
  pages={7252--7273},
  year={2025}
}

@article{wang2025wilddoc,
  title={WildDoc: How far are we from achieving comprehensive and robust document understanding in the wild?},
  author={Wang, Anlong and Tang, Jingqun and Liao, Lei and Feng, Hao and Liu, Qi and Fei, Xingjian and Lu, Jinghui and Wang, Han and Liu, Hao and Liu, Yuliang and others},
  journal={Proceedings of the 2025 Conference on Empirical Methods in Natural Language Processing},
  year={2025}
}

@article{shan2024mctbench,
  title={MCTBench: Multimodal cognition towards text-rich visual scenes benchmark},
  author={Shan, Biluo and Fei, Xingjian and Shi, Wenqiang and Wang, Anlong and Tang, Guozhi and Liao, Lei and Tang, Jingqun and Bai, Xiang and Huang, Can},
  journal={arXiv preprint arXiv:2410.11538},
  year={2024}
}

@inproceedings{wang2024pargo,
  title={PARGO: Bridging vision-language with partial and global views},
  author={Wang, Anlong and Shan, Biluo and Shi, Wenqiang and Lin, Kun-Yu and Fei, Xingjian and Tang, Guozhi and Liao, Lei and Tang, Jingqun and Huang, Can and others},
  booktitle={Proceedings of the AAAI Conference on Artificial Intelligence},
  year={2024}
}

@article{feng2026dolphin_v2,
  title={Dolphin-v2: Universal document parsing via scalable anchor prompting},
  author={Feng, Hao and Shi, Wenqiang and Zhang, Kaijie and Fei, Xingjian and Liao, Lei and Yang, Dingkun and Du, Yibo and Wu, Xiao and Tang, Jingqun and Liu, Yuliang and others},
  journal={arXiv preprint arXiv:2602.05384},
  year={2026}
}

@article{wang2025vision,
  title={Vision as LoRA},
  author={Wang, Han and Ye, Yongjie and Li, Bingqi and Nie, Yuhang and Lu, Jinghui and Tang, Jingqun and Wang, Yanjie and Huang, Can},
  journal={arXiv preprint arXiv:2503.20680},
  year={2025}
}

@inproceedings{sun2024attentive,
  title={Attentive Eraser: Unleashing diffusion model's object removal potential via self-attention redirection guidance},
  author={Sun, Wenhao and Cui, Benlei and Tang, Jingqun and Dong, Xue-Mei},
  booktitle={Proceedings of the AAAI Conference on Artificial Intelligence},
  year={2024}
}

@article{fei2025advancing,
  title={Advancing sequential numerical prediction in autoregressive models},
  author={Fei, Xingjian and Lu, Jinghui and Sun, Qian and Feng, Hao and Wang, Yanjie and Shi, Wenqiang and Wang, Anlong and Tang, Jingqun and Huang, Can},
  journal={Proceedings of the 63rd Annual Meeting of the Association for Computational Linguistics},
  year={2025}
}

@article{zhu2026textpecker,
  title={TextPecker: Rewarding structural anomaly quantification for enhancing visual text rendering},
  author={Zhu, Hao and Liu, Yuliang and Wu, Xiao and Wang, Anlong and Feng, Hao and Yang, Dingkun and Feng, Chen and Huang, Can and Tang, Jingqun and others},
  journal={arXiv preprint arXiv:2602.20903},
  year={2026}
}
}

\end{document}